%% file: main_v2.tex
\documentclass[conference]{IEEEtran}
\IEEEoverridecommandlockouts

\usepackage{rotating}
\usepackage{cite}
\usepackage{amsmath,amssymb,amsfonts}
\usepackage{algorithmic}
\usepackage{graphicx}
\usepackage{textcomp}
\usepackage{pgfplots}
\usepackage{multirow}
\usepackage{comment}
\usepackage{tikz}
\usetikzlibrary{matrix}
\usepackage{tablefootnote}

\newcommand{\agents}{MLLM agents}
\newcommand{\agent}{MLLM agent}
\newcommand{\QwenVL}{Qwen3-VL:8B}

\usepackage{pgfplots}
\pgfplotsset{compat=1.18} 

\usepackage{bm}
\makeatletter
\AtBeginDocument{\DeclareMathVersion{bold}
\SetSymbolFont{operators}{bold}{T1}{times}{b}{n}
\DeclareSymbolFont{NewLetters}{T1}{times}{b}{it}
\SetSymbolFont{NewLetters}{bold}{T1}{times}{b}{it}
\SetMathAlphabet{\mathrm}{bold}{T1}{times}{b}{n}
\SetMathAlphabet{\mathit}{bold}{T1}{times}{b}{it}
\SetMathAlphabet{\mathbf}{bold}{T1}{times}{b}{n}
\SetMathAlphabet{\mathtt}{bold}{OT1}{pcr}{b}{n}
\SetSymbolFont{symbols}{bold}{OMS}{cmsy}{b}{n}
\renewcommand\boldmath{\@nomath\boldmath\mathversion{bold}}}
\makeatother

\usepackage{adjustbox}
\usepackage{url}

\usepackage{textcomp}
\usepackage{booktabs}
\usepackage{makecell}
\usepackage{multirow}
\usepackage{tikz}
\usepackage{pgfplots}
\usetikzlibrary{calc}
\usepackage {color, graphicx}
\usetikzlibrary{decorations.pathmorphing}
\usetikzlibrary {patterns,shapes,shadows,decorations.pathreplacing}
\usepackage{amssymb}
\usetikzlibrary{arrows,shadows,positioning}
\usetikzlibrary{arrows.meta}
\usetikzlibrary{snakes}

\usepackage{ifthen}
\usepackage{amssymb}
\usepackage{pifont}
\usetikzlibrary{shapes.symbols}
\usepackage{graphicx,subfig}
\usepackage{color, colortbl}	
\usepackage{lineno}
\usepgfplotslibrary{groupplots}
\pgfplotsset{compat=1.18}

\usepackage{algorithm}
\usepackage{algorithmic}
\usepackage{multirow}
\usepackage{tikz}
\usepackage{graphicx}
\usepackage{float}

\definecolor{pal1}{HTML}{FEEAC9}
\definecolor{pastGreen}{HTML}{C1E1C1}     
\definecolor{pastLGreen}{HTML}{E2F0CB}    
\definecolor{pastYellow}{HTML}{FDFD96}    
\definecolor{pastLRed}{HTML}{FFB7B2}      
\definecolor{pastRed}{HTML}{FF6961}       

\usepackage{forest}

\usetikzlibrary{positioning} 

\usetikzlibrary{shapes.geometric, arrows, positioning}

\tikzstyle{startstop} = [rectangle, rounded corners, minimum width=3cm, minimum height=1cm, text centered, draw=black, fill=gray!20]
\tikzstyle{process} = [rectangle, minimum width=3cm, minimum height=1cm, text centered, draw=black, fill=blue!20]
\tikzstyle{decision} = [diamond, minimum width=3cm, minimum height=1cm, text centered, draw=black, fill=green!20]
\tikzstyle{arrow} = [thick,->,>=stealth]

\usepackage{longtable}
\usepackage[x11names]{xcolor}  

\def\BibTeX{{\rm B\kern-.05em{\sc i\kern-.025em b}\kern-.08em
    T\kern-.1667em\lower.7ex\hbox{E}\kern-.125emX}}
\begin{document}

\title{Stable Behavior, Limited Variation: Persona Validity in LLM Agents for Urban Sentiment Perception}

\author{\IEEEauthorblockN{Neemias B da Silva\IEEEauthorrefmark{1}, Rodrigo Minetto\IEEEauthorrefmark{1}, Daniel Silver\IEEEauthorrefmark{2},Thiago H Silva\IEEEauthorrefmark{1}\IEEEauthorrefmark{2}}
\IEEEauthorblockA{\IEEEauthorrefmark{1}Universidade Tecnol\'{o}gica Federal do Paran\'a (UTFPR), Curitiba, Brazil}
\IEEEauthorblockA{\IEEEauthorrefmark{2}University of Toronto, Toronto, Canada\\
Email: neemiasbuceli@alunos.utfpr.edu.br, rminetto@utfpr.edu.br, \{dan.silver,th.silva\}@utoronto.ca}
}
\maketitle

\begin{abstract}
Large Language Models (LLMs) are increasingly used as proxies for human perception in urban analysis, yet it remains unclear whether persona prompting produces meaningful and reproducible behavioral variation. We investigate whether distinct personas influence urban sentiment judgments generated by multimodal LLMs. Using a factorial set of personas spanning gender, economic status, political orientation, and personality, we instantiate multiple agents per persona to evaluate urban scene images from the PerceptSent dataset and assess within-persona convergence and cross-persona differentiation. Results show strong convergence among agents sharing a persona, indicating stable and reproducible behavior. However, cross-persona differentiation is limited: economic status and personality induce statistically detectable but practically modest variation, while gender shows no measurable effect and political orientation only negligible impact. Agents also exhibit an extremity bias, collapsing intermediate sentiment categories common in human annotations. As a result, performance remains strong on coarse-grained polarity tasks but degrades as sentiment resolution increases, suggesting that persona prompting does not capture fine-grained perceptual judgments. To isolate the contribution of persona conditioning, we additionally evaluate the same model without personas. Surprisingly, the no-persona model matches or exceeds persona-conditioned agreement with human labels across all task variants, suggesting that persona prompting may add limited annotation value in this setting.\end{abstract}

\begin{IEEEkeywords}
Persona Prompting,
Urban Perception,
Sentiment Analysis,
Human-AI Alignment

\end{IEEEkeywords}

\section{Introduction}

Large language models (LLMs) are increasingly used as simulated agents to approximate human judgments \cite{Argyle_Busby_Fulda_Gubler_Rytting_Wingate_2023,aherICML2023,park2024generativeagentsimulations1000}, yet it remains unclear whether they can reproduce not only consistent but also sufficiently diverse judgments \cite{beck-etal-2024-sensitivity,hu-collier-2024-quantifying,li2025simulating}. A common mechanism for introducing diversity in LLM agents is \emph{attribute-based persona prompting}, in which attributes such as gender, economic status, personality, and political orientation are injected into prompts to guide model behavior~\cite{Argyle_Busby_Fulda_Gubler_Rytting_Wingate_2023,hu-collier-2024-quantifying, Bail2024PNAS}. Despite its growing adoption, a central question remains: \emph{Do personas produce distinct, reproducible, and sufficiently granular patterns of human perceptual judgment, or merely create the appearance of diversity without substantial behavioral differentiation?} Addressing this question is critical for evaluating the validity of LLM-based agent simulations, particularly in light of concerns about the reduction of within-group variance~\cite{wang2025large}, the lack of alignment with human psychometric baselines~\cite{wang_zou_yan_guo_sun_xiao_zhang_2024}, and output variance across prompt templates~\cite{lutz-etal-2025-prompt}.

This question is especially relevant for perceptual tasks in urban environments. Judgments about urban scenes---such as whether a place appears pleasant, unsafe, or neglected---vary across individuals and social contexts~\cite{PerceptSent}. Prior work on LLM-based social simulation and persona prompting suggests that demographic and psychological attributes can influence model behavior and are commonly used to define synthetic agents~\cite{Argyle_Busby_Fulda_Gubler_Rytting_Wingate_2023, hu-collier-2024-quantifying}. Motivated by this, we adopt gender, economic status, political orientation, and personality as controlled persona attributes to evaluate whether such prompts induce stable and distinguishable patterns of urban sentiment judgment.

Collecting annotations that reflect this diversity typically requires large numbers of human annotators, making dataset construction costly and time-consuming~\cite{10.1145/3385186}. Multimodal LLMs (MLLMs) offer a potential alternative as \emph{synthetic annotators}, capable of interpreting images and generating structured perceptual descriptions. However, their usefulness depends not only on producing stable outputs, but also on capturing variability at the granularity of human perception, particularly the ability to represent intermediate and ambiguous judgments.

In this paper, we empirically evaluate whether attribute-based personas induce distinguishable and stable behavior in MLLM agents performing urban image sentiment annotation. We construct a factorial design of 24 personas combining gender, economic status, political orientation, and personality traits, and instantiate 50 independent agents per persona to annotate 50 urban scene images from the PerceptSent dataset~\cite{PerceptSent}. We analyze both \emph{within-persona convergence} (consistency among agents sharing a persona) and \emph{cross-persona differentiation} (differences across personas).

Our results show strong within-persona convergence, indicating stable and reproducible behavior. However, cross-persona differentiation shows complex patterns: economic status and personality show statistically detectable but practically modest variation in annotations, while gender shows no measurable effect; political orientation has only a negligible impact. Moreover, agents exhibit a systematic extremity bias, collapsing intermediate sentiment categories that are prevalent in human annotations. As a result, while performance remains strong on coarse-grained polarity judgments, it degrades as task resolution increases. These findings suggest that simple attribute-based persona prompting stabilizes outputs but fails to generate sufficiently fine-grained behavioral variation, limiting its ability to capture the fine-grained perceptual judgments characteristic of human labelers.

This paper makes four main contributions:

\begin{itemize}
    \item \textbf{A framework for evaluating persona validity in MLLM agents.} We propose a methodology that jointly measures within-persona convergence and cross-persona differentiation to assess whether persona prompts produce meaningful behavioral differences.
    
    \item \textbf{A controlled large-scale study of persona profiles in multimodal annotation.} We instantiate 1{,}200 persona-conditioned agents across 24 persona profiles to evaluate 50 urban scene images, analyzing sentiment predictions under a controlled factorial design. Also, we share code and generated data publicly\footnote{\url{https://github.com/neemiasbsilva/mllm-persona-evaluation}.}.
    
    \item \textbf{Insights into the behavioral and resolution-dependent effects of persona prompting.} Our results show that economic status and personality drive detectable but limited cross-persona differentiation, while gender shows no measurable effect, and political orientation has only a negligible impact. More broadly, we identify a persistent tendency in MLLM agent judgments: while persona prompting stabilizes outputs and supports coarse-grained polarity judgments, it fails to produce fine-grained variation, collapsing intermediate sentiment categories and limiting its ability to capture fine-grained perceptual judgments. This reveals a resolution-dependent limitation of persona prompting.

    \item \textbf{A no-persona ablation suggests that persona conditioning may not improve annotation quality.} We run the same model on the same 50 images without any persona prompt, under two reasoning modes (with and without extended chain-of-thought (CoT)), with 5 independent repetitions per condition. The no-persona \agents{} achieve agreement comparable to, and in some cases higher than, the 1{,}200 persona-conditioned \agents{}. Furthermore, no-persona predictions show a substantially less extreme bimodal collapse, with Neutral class usage close to human rate (23--25\% vs.\ 21\%).
\end{itemize}

The remainder of this paper is organized as follows. Section~\ref{sec:related} reviews related work. Section~\ref{sec:method} describes the methodology. Section~\ref{sec:results} presents experimental results. Section~\ref{secLimiEth} discusses limitations and ethical considerations. Section~\ref{sec:conclusion} concludes the paper.

\section{Related Work}\label{sec:related}

Recent advances in LLMs have motivated their use as computational
proxies for human behavior, with studies showing they can reproduce
stylized experimental findings and support survey research, behavioral simulation, and content analysis
\cite{Apostolos2024,aherICML2023,Bail2024PNAS}. At the same time,
prior work highlights important limitations, including bias, reduced human variability, and challenges for psychometric validity and reproducibility
\cite{wang_zou_yan_guo_sun_xiao_zhang_2024,wang2025large}. While some studies find partial alignment between LLMs and human judgments in specific tasks \cite{de2025humans}, it remains unclear whether LLM-based agents can reliably substitute for human participants.

More closely related is work on attribute-based persona prompting, where
demographic or psychological attributes guide model behavior. Prior
studies show that personas can affect outputs
\cite{lutz-etal-2025-prompt,malik-etal-2025-llms}, while raising
concerns that synthetic personas may distort underlying belief
structures \cite{barrie_cerina_2026}. Previous work mainly evaluates cross-group differences in text generation tasks, with limited attention to within-persona stability, meaningful cross-persona differentiation, or improvement in agreement with human labels relative to a neutral prompt, particularly in multimodal perceptual tasks. Work specifically on MLLMs as synthetic annotators for perceptual and affective tasks remains sparse, with most studies focusing on text-only models \cite{10.1145/3385186,PerceptSent,dasilva2025multimodalllmssentiment}. Addressing these gaps is the focus of this work.

\section{Methodology}\label{sec:method}

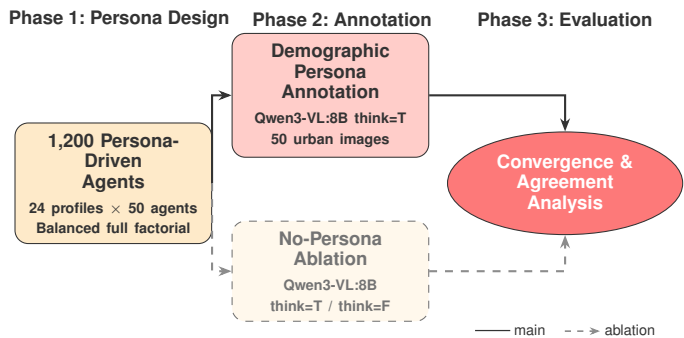
\begin{figure}[htb]
    \centering
    \input{diagrams/methodology_overview_v2}
    \caption{
        High-level methodology overview.
    }
    \label{fig:methodology_overview}
\end{figure}

The methodology comprises three phases (Figure~\ref{fig:methodology_overview}): (i)~a \emph{persona design} constructs a balanced factorial set of attribute-based personas; (ii)~an \emph{annotation} phase that runs two parallel conditions on the same 50 urban scene images: the main \emph{persona} condition, in which 1{,}200 persona-conditioned agents annotate each image, and a \emph{no-persona ablation}, in which the same model annotates each image with a neutral observer prompt under two reasoning modes; and (iii)~an \emph{evaluation} phase that measures within-persona convergence and agreement with human ground truth across all conditions.

\subsection{Persona Design}
\label{sec:method:personas}

We construct a balanced full factorial design across four persona attributes (Table~\ref{tab:dimensions}). 
We selected three commonly studied sociodemographic attributes (gender, economic status, and political orientation) and included personality as an additional attribute, following the suggestion that persona effects may depend not only on demographic cues but also on psychological traits that shape variability in agent behavior \cite{hu-collier-2024-quantifying}. These archetypes were chosen to span contrasting cognitive and affective orientations. The number of levels per attribute was kept small to maintain a tractable full factorial design and keep the experimental workload computationally feasible.

\begin{table}[h]
    \centering
    \scriptsize
    \caption{Persona attributes and category levels.}
    \begin{tabular}{c|c|c}
         \toprule
        \textbf{Attribute}           & \textbf{Categories}                        & \textbf{Levels} \\
        \midrule
        Gender              & Male, Female                      & 2 \\
        Economic Status     & Low income, High income           & 2 \\
        Political Spectrum  & Progressive, Conservative         & 2 \\
        Personality         & Analytical, Empathetic, Pragmatic  & 3 \\
        \bottomrule
    \end{tabular}
    \label{tab:dimensions}
\end{table}

This design yields $2 \times 2 \times 2 \times 3 = 24$ unique personas. Each persona is instantiated with 50 independent agents, resulting in 1{,}200 agents in total. Agents are generated independently from the same persona prompt without shared context or conversation history. Residual variation within a persona arises only from low-temperature ($0.1$) floating-point non-determinism.

Each persona is defined by a set of four attributes injected into a structured system prompt. For example: \emph{Gender: Male; Economic status: Low income; Political spectrum: Progressive; Personality archetype: Analytical}.

\subsection{Image Dataset}\label{sec:method:dataset}

The \textit{PerceptSent} dataset~\cite{PerceptSent} contains 5{,}000 urban scene photographs, each annotated by five human workers with: (i)~a sentiment score on a 5-point ordinal scale (Negative, Slightly Negative, Neutral, Slightly Positive, Positive); and (ii)~free-text perception labels drawn from a closed vocabulary.

\begin{figure}[h]
  \centering
  \subfloat[Human modal sentiment]{\includegraphics[width=0.18\textwidth, keepaspectratio]{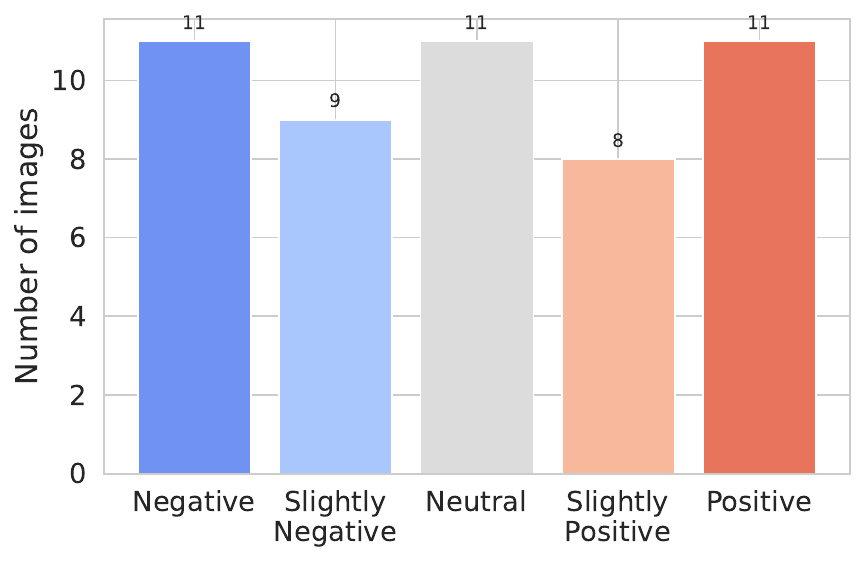}}
  \hfill
  \subfloat[Scene Type]{\includegraphics[width=0.10\textwidth, keepaspectratio]{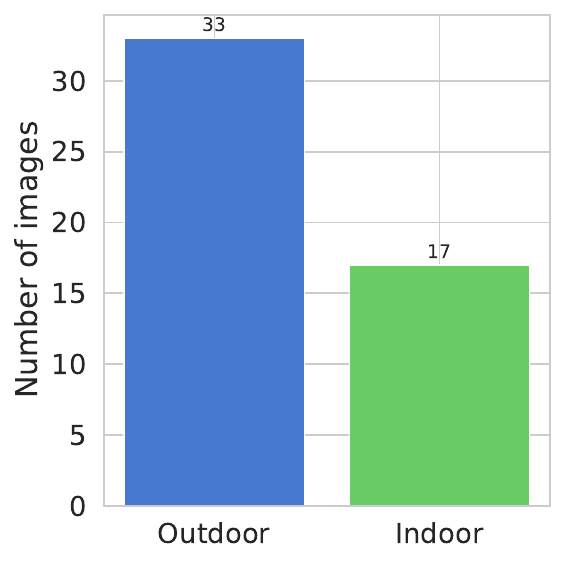}}
  \hfill
  \subfloat[Sentiment $\times$ scene type (image count)]{\includegraphics[width=0.20\textwidth, keepaspectratio]{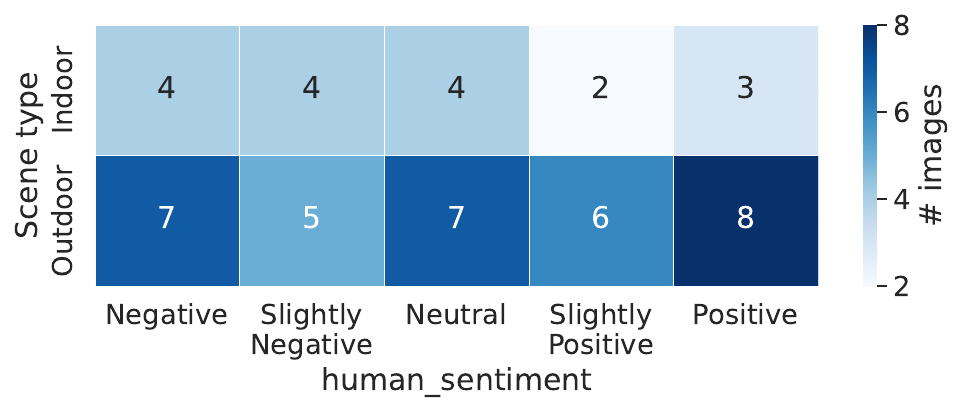}}
  \caption{
    Characteristics of the studied dataset.}
  \label{fig:img_dist}
\end{figure}

A stratified subset of 50 images is drawn proportional to $\text{sentiment} \times \frac{indoor}{outdoor}$ to ensure balanced coverage of both affective polarity and spatial context. Figure~\ref{fig:img_dist} shows details of the selected sample. The human modal sentiment distribution of the subset is approximately uniform across the five classes (8--12 images per class), with a slight over-representation of Negative scenes consistent with the full dataset's human label distribution. The indoor/outdoor split in the subset reflects the strong compositional skew of the source dataset: 92.0\% of the 5{,}000 PerceptSent images depict outdoor scenes (4{,}598 outdoor vs.\ 402 indoor), a bias inherent to how urban scene collections are typically assembled (e.g., street, views, public spaces, buildings). Stratified sampling, therefore, preserves this ratio rather than artificially balancing it, ensuring the subset remains representative of the population from which conclusions will be drawn. The experiment targets $1{,}200 \times 50 = 60{,}000$ annotation triples.

\subsection{Annotation Pipeline}\label{sec:method:annotation}

The annotation pipeline is implemented as a three-stage \emph{LangGraph}\footnote{https://www.langchain.com/langgraph} workflow with conditional retry logic (Figure~\ref{fig:annotation_pipeline}).

\begin{figure}[htb]
  \centering
  \resizebox{0.50\textwidth}{!}{%
      \input{diagrams/annotation_pipeline_redopt}
  }
  \caption{Three-stage annotation pipeline comprising image loading, prompt assembly, and vision annotation. The Eval \& Retry cluster handles timeout, JSON parsing, and label errors with up to 3 retries before logging failures.
    }
  \label{fig:annotation_pipeline}
\end{figure}

\subsubsection{Model Configuration}

All annotations use \emph{{\QwenVL}} and are run locally via Ollama. See Table~\ref{tab:model_config} for configuration details.

\begin{table}[h]
  \centering
    \scriptsize
  \caption{Vision model configuration.}
  \label{tab:model_config}
  \begin{tabular}{c|c}
    \toprule
   \textbf{Parameter} & \textbf{Value} \\
    \midrule
    Model            & \texttt{\QwenVL{}} \\
    Temperature      & 0.1 \\
    Chain-of-thought & Enabled (\texttt{think=True}\tablefootnote{\texttt{think=False} was also evaluated in the no-persona ablation condition (Section \ref{sec:method:nopersona})}) \\
    Seed             & 42 \\
    Context window   & 4{,}096 tokens \\
    Max output tokens & Unlimited ($-1$) \\
    Timeout          & 360\,s per call \\
    \bottomrule
  \end{tabular}
\end{table}

\subsubsection{Pipeline Nodes}

\textit{Image Loader} reads JPEG images from local storage and encodes them as base64 strings. Images are pre-cached in memory before the annotation loop to avoid repeated disk I/O.

\textit{Prompt Assembler} constructs the system prompt by injecting the four attribute tags into the following template:

\begin{quote}\tiny\ttfamily
    You are a person with the following persona profile:\\
        \ \ - Gender: \{gender\}\\
        \ \ - Economic status: \{economic\_status\}\\
        \ \ - Political leaning: \{political\_spectrum\}\\
        \ \ - Personality archetype: \{personality\}\\[4pt]
    CRITICAL RULES:\\
        \ \ 1. You are this person. You are NOT an AI.\\
        \ \ 2. Evaluate the image exactly as this person would.\\
        \ \ 3. Do not break character.\\[4pt]
    TASK --- IMAGE SENTIMENT ANNOTATION: Look at the image and respond with a JSON object and nothing else.\\[4pt]
    The JSON must have exactly these four keys:\\
        \ \ \texttt{"sentiment"} : one of \{Negative, SlightlyNegative, Neutral, SlightlyPositive, Positive\}\\
        \ \ \texttt{"perceptions"} : a JSON array of 1--5 labels chosen EXCLUSIVELY from the PERCEPTION VOCABULARY listed below --- do NOT invent new labels\\
        \ \ \texttt{"caption"} : a 1--2 sentence objective description of what is shown in the image, written independently of your persona\\
        \ \ \texttt{"justification"} : a single sentence in YOUR voice explaining your sentiment score\\[4pt]
    PERCEPTION VOCABULARY (593 labels --- choose 1--5 that best match the image).\\[4pt]
    Example response: \{"sentiment": "Negative", "perceptions": ["Violence", "Debris/ Destruction"], "caption": "A damaged street with rubble and broken glass scattered across the pavement.",\\
    \ "justification": "Looks like the same mess they always leave after those marches downtown."\}\\[4pt]
    Return ONLY the JSON object. No markdown, no code fences, no additional text.
\end{quote}

The annotation instruction needs a JSON response with the field \texttt{sentiment}, which represents a 5-class ordinal.

\textit{Vision Annotator} sends the assembled prompt and base64 image to the \emph{\QwenVL} model via the Ollama API. A multi-layer error-handling strategy with up to \textit{3 parse retries}: (i)~timeouts, retried with a compact JSON-only prompt; (ii)~JSON parse errors, repaired via regex extraction and correction instructions; (iii)~label normalization via fuzzy matching. Annotations exhausting all retries are logged and skipped.

\subsubsection{Evaluate Response}

The pipeline uses semaphore-bounded async concurrency to evaluate the response. On startup, existing \emph{JSONL} output files are read to skip completed \texttt{(persona\_id, image\_id)} pairs, enabling fault-tolerant resumption after crashes. Successful responses and failed annotations are written to separate JSONL files, allowing concurrent runs of different conditions without file locking conflicts.

\subsection{No-Persona Ablation Condition}\label{sec:method:nopersona}

To isolate the contribution of persona conditioning, we run the same \emph{\QwenVL{}} on the same 50 images without injecting any persona. The system prompt is replaced with a neutral observer instruction:

\begin{quote}\tiny\ttfamily
    You are an impartial image annotator. Evaluate each image objectively,\\
    without any personal or demographic lens.\\[4pt]
    [Same JSON annotation task and perception vocabulary as above.]
\end{quote}

We evaluate two reasoning modes: \texttt{think=True} (extended CoT, identical to the persona setting) and \texttt{think=False} (direct response without an internal reasoning step), each run 5 independent times over the same 50 images (250 annotations per condition). The modal sentiment across the 5 runs is used as the condition's prediction for each image. These two conditions let us ask, independently, whether persona prompting adds annotation value beyond the model's intrinsic visual understanding, and whether CoT reasoning influences agreement with human labels when no persona is present.

\subsection{Evaluation Metrics}\label{sec:method:metrics}

We assess \emph{within-persona convergence} using two measures. The \emph{modal ratio} is the fraction of annotations within each (\texttt{persona\_id}, \texttt{image\_id}) group that match the modal sentiment label. A value of 1.0 indicates full consensus; the chance baseline for 5 classes is 0.2. The \emph{sentiment variance} is computed over integer-encoded scores ($\text{Neg}=-2$ to $\text{Pos}=+2$) within each group. A value of 0.0 indicates that all {\agents} assigned the same class, while higher values reflect greater dispersion.

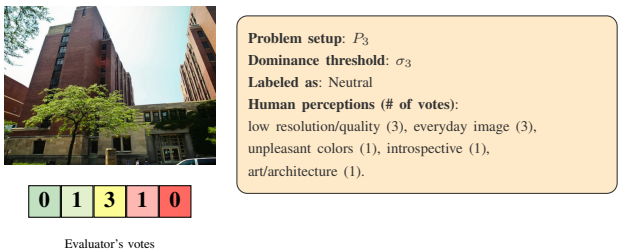
\begin{figure}[!h]
    \centering
    \begin{tikzpicture}[
        scale=0.65, 
        every node/.style={scale=0.70},
        annotbox/.style={
            draw=black!80, 
            fill=pal1, 
            rounded corners=4pt, 
            text width=6.8cm, 
            inner sep=6pt, 
            align=left, 
            text=black!80
        }
    ]
        \draw(0.0,3.0) node[] (img1) {\includegraphics[height=3cm,width=4cm]{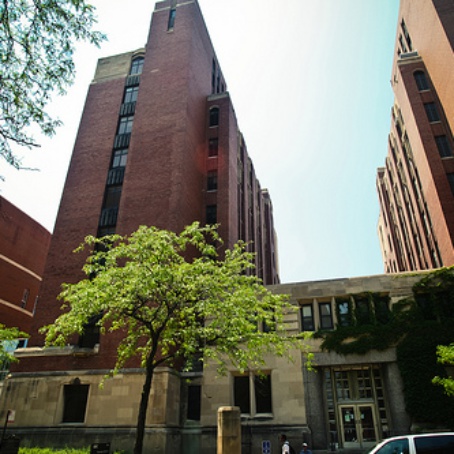}};

        \matrix(m) [
            matrix of math nodes,  
            nodes in empty cells, 
            nodes={draw, minimum size=0.6cm, outer sep=0pt, text height=1.5ex, text depth=.15ex},
            below=2pt of img1
        ] 
        {
            |[draw,fill=pastGreen]| \large \textbf{0} &  
            |[draw,fill=pastLGreen]|  \large \textbf{1} & 
            |[draw,fill=pastYellow]| \large \textbf{3} & 
            |[draw,fill=pastLRed]| \large \textbf{1} & 
            |[draw,fill=pastRed]| \large \textbf{0}\\
        };

        \draw node[below=2pt of m] (text0) {\scriptsize Evaluator's votes};

        \node[annotbox] (box1) at ([xshift=4.2cm, yshift=-0.4cm]img1.east) {%
            \begin{tabular}{@{}l@{}}
                \footnotesize \textbf{Problem setup}: $P_3$ \\
                \footnotesize \textbf{Dominance threshold}: $\sigma_3$\\
                \footnotesize \textbf{Labeled as}: Neutral\\
                \footnotesize \textbf{Human perceptions (\# of votes)}:\\ 
                \footnotesize low resolution/quality (3), everyday image (3),\\
                \footnotesize unpleasant colors (1), introspective (1), \\
                \footnotesize art/architecture (1).
            \end{tabular}
        };
        
    \end{tikzpicture}
    \caption{PerceptSent dataset example.}
    \label{fig:percept_example}
\end{figure}

Agreement with human ground truth is evaluated using 5-fold cross-validation over the Cartesian product of consensus thresholds $\sigma$ and task formulations $P$. Figure~\ref{fig:percept_example} exemplifies how one particular image is annotated and why this particular image has been assigned for $P_3$ and $\sigma_3$ agreement. Here, $\sigma \in \{3,4,5\}$ denotes the minimum number of human annotators required to agree (higher $\sigma$ yields stricter consensus), and $P \in \{\text{$P_2^{-}$, $P_2^{+}$, $P_3$, $P_5$}\}$ defines the task: binary negative polarity ($P_2^{-}$), binary positive polarity ($P_2^{+}$), 3-class ($P_3$), and 5-class ($P_5$).

Since the annotation pipeline performs no model training or fine-tuning, the same model weights are used for every agent call. All images that satisfy the $\sigma$ criterion are included in every fold. To estimate variability, we resample 60\% of the annotation pool per image in each fold (716 annotations per image, from $\approx$1{,}194 total), recompute the modal sentiment, and evaluate performance. This yields between 2{,}868 and 35{,}850 annotations per fold, depending on $\sigma$. Metrics are averaged across folds.

Uncertainty is reported as 95\% confidence intervals. This design progressively filters ambiguous images, ensuring that evaluation focuses on cases with stronger human agreement.

We report Macro~F1 (primary metric, following the MLLMsent framework~\cite{dasilva2025multimodalllmssentiment}) and Cohen’s~$\kappa$ (unweighted for binary tasks, linear for 3-class, and quadratic for 5-class).

\section{Experiments and Results}\label{sec:results}

\subsection{Annotation Statistics}\label{sec:results:stats}

The annotation run produced \textit{59{,}708 successful annotations} from 1{,}200 unique agents across 50 images, covering all 24 persona profiles. Table~\ref{tab:annot_stats} summarizes the pipeline statistics. The failure rate is 0.49\% (292 annotations), with 98.2\% of successful annotations completed on the first validation attempt. The no-persona ablation runs each produced exactly 250 annotations across the same 50 images with zero failures, yielding one annotation per image per condition.

\begin{table}[h]
  \centering
  \scriptsize
  \caption{Annotation pipeline statistics across all three conditions.}
  \label{tab:annot_stats}
  \begin{tabular}{c|c|c|c}
    \toprule
    \textbf{Metric} & \textbf{Persona} & \textbf{No-pers.} & \textbf{No-pers.} \\
                    &                   & \textbf{think=T}  & \textbf{think=F} \\
    \midrule
    Successful ann.       & 59{,}708 & 250   & 250   \\
    Failed ann.           & 292      & 0    & 0    \\
    Failure rate          & 0.49\%   & 0\%  & 0\%  \\
    Unique agents         & 1{,}200  & 5   & 5    \\
    Personas / conditions & 24       & N/A  & N/A  \\
    Unique images         & 50       & 50   & 50   \\
    Ann.\ / image         & $\approx$1{,}194 & 5 & 5 \\
    \bottomrule
  \end{tabular}
\end{table}

Failures are demographically balanced (50.6\% Male, 49.4\% Female; 50.6\% Low income, 49.4\% High income) and concentrate on a small number of structurally complex scenes, ruling out {\agents}-specific parsing effects and indicating that errors are primarily content-driven.

\subsection{Predicted Sentiment Distribution}\label{sec:results:dist}

\begin{figure}[h]
  \centering
  \includegraphics[width=0.49\textwidth, keepaspectratio]{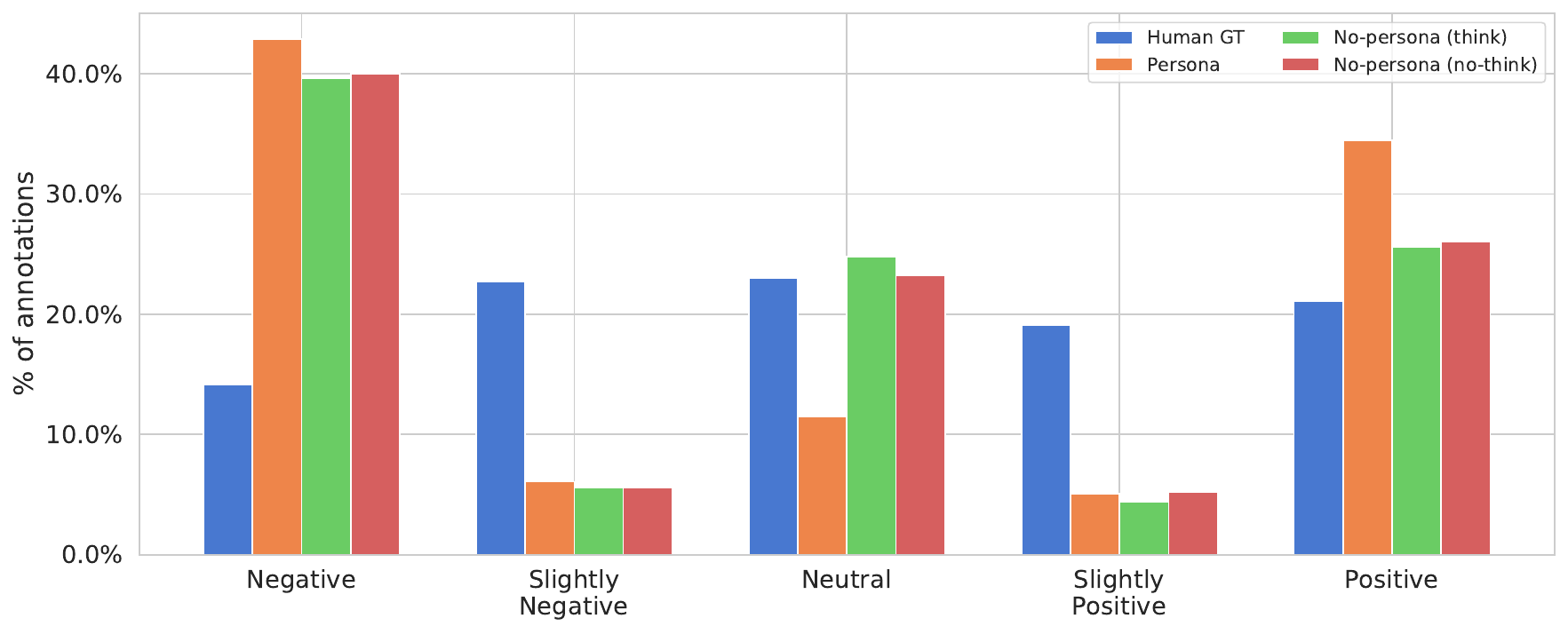}
  \caption{
        Predicted sentiment distribution across all three conditions vs.\ human ground truth (GT). 
    }
  \label{fig:sent_dist}
\end{figure}

Figure~\ref{fig:sent_dist} compares predicted sentiment distributions across all three conditions against the human ground truth. Persona prompting exhibits a pronounced bimodal collapse, with 77.8\% of predictions concentrated at the polarity extremes (Negative + Positive), compared to 35.3\% in human annotations. Intermediate categories are correspondingly underused, indicating an extremity bias relative to the central tendency commonly observed in human responses \cite{tourangeau1996asking}.

By contrast, the no-persona conditions are substantially better
calibrated, reducing extreme-label usage (65--66\%) and producing
Neutral rates (23--25\%) close to the human baseline (21\%). This
suggests persona conditioning may amplify rather than mitigate the
bimodal collapse. Importantly, disagreement arises primarily from
collapsed intermediate categories rather than polarity inversions:
errors are largely adjacent in the 5-class task (Section
\ref{sec:results:gt}), indicating a failure of granularity rather than scene comprehension.

\subsection{Within-Persona Sentiment Convergence}\label{sec:result:convergence}

{\agents} instantiated with the same persona show strong within-persona convergence. For each of the $24\times50\!=\!1{,}200$ (\texttt{persona\_id}, \texttt{image\_id}) groups, we compute two complementary metrics: \emph{modal ratio} and \emph{sentiment variance}. Table~\ref{tab:convergence} summarizes these results across all groups.

\begin{table}[h]
  \centering
  \scriptsize
  \caption{Within-persona convergence metrics. \emph{Agents/group}: number of agents in each group.}
  \label{tab:convergence}
  \begin{tabular}{l|l|l|l|l}
    \toprule
    \textbf{Metric} & \textbf{Mean} & \textbf{$\pm$ SD} & \textbf{Q1} & \textbf{Median} \\
    \midrule
    Modal ratio        & 0.871 & $\pm$\,0.185 & 0.760 & 0.980 \\
    Sentiment variance & 0.211 & $\pm$\,0.405 & 0.000 & 0.000 \\
    Agents/group   & 49.6  & $\pm$\,1.4   & 50    & 50    \\
    \bottomrule
  \end{tabular}
\end{table}

A mean modal ratio of 0.871 and median of 0.980 (with median sentiment variance of 0.000) indicate near-unanimous agreement within most groups, many of which achieve full consensus ($\text{modal ratio}=1.0$). Although a minority of groups show disagreement, the vast majority exhibit full or near-full consensus, confirming stable and reproducible sentiment judgments across identically prompted {\agents}. This high convergence is consistent with the near-deterministic setup (temperature~0.1, seed~42, fixed model weights and context window), with residual disagreement likely reflecting floating-point non-determinism.

\subsection{Do Personas Produce Distinct Judgments? Cross-Persona Differentiation}\label{sec:results:variation}

Establishing convergence within personas is only half of the answer; the other is whether \emph{different} personas produce systematically different sentiments. Figure~\ref{fig:profile_heatmap} shows the row-normalized sentiment distribution for all 24 personas. Some personas are more Negative-biased (top), while others are more Positive-biased (bottom). Notably, intermediate classes (Neutral, Slightly Negative, Slightly Positive) are largely absent across all personas, a direct consequence of the bimodal collapse noted above.

Despite this, the \emph{relative ordering} of Negative-to-Positive proportions varies meaningfully across personas, indicating that persona composition modulates the {\agents}' sentiment prior. A Kruskal--Wallis test over ordinal sentiment scores ($[-2, +2]$) grouped by the 24 personas rejects the null hypothesis of equal distributions ($H(23)\!=\!489.52$, $p\!=\!5.33\times10^{-89}$, $\varepsilon^2\!=\!0.0078$), indicating statistically significant differences. However, effect sizes are small, suggesting limited practical differentiation.

\begin{figure}[t]
    \centering
  \includegraphics[width=0.5\textwidth]{figures/fig_rq1a_profile_sentiment_heatmap.pdf}
  \caption{
    Row-normalized sentiment proportion heatmap across all 24 persona profiles, sorted by descending Negative fraction. Each cell shows the fraction of annotations in a given sentiment class. }
  \label{fig:profile_heatmap}
\end{figure}

\begin{figure}[tbh]
  \centering \includegraphics[width=.5\textwidth]{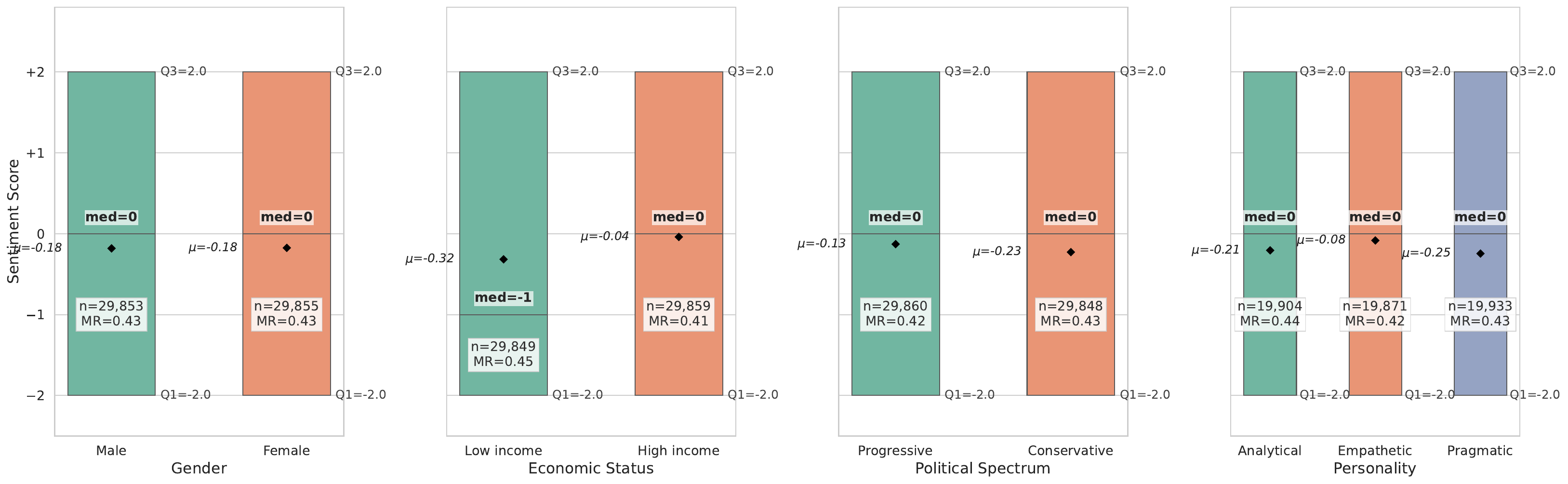}
  \caption{
    Sentiment distribution by persona attribute ($[-2,+2]$ scale).}
  \label{fig:boxplot_dim}
\end{figure}

To identify which persona attributes drive this variation, we decompose sentiment distributions along each attribute (Figure~\ref{fig:boxplot_dim}). Factor-level Mann--Whitney U and Kruskal--Wallis tests over ordinal scores reveal which attributes produce statistically significant separation.

Economic status shows the largest effect: High-income personas have mean $\mu\!=\!{-0.04}$ (median $=0$), while Low-income personas have $\mu\!=\!{-0.32}$ (median $=-1$), making it the only attribute that shifts the median below zero—indicating a qualitative polarity shift  ($U\!=\!482{,}309{,}610$, $p\!=\!4.29\times10^{-77}$, $r\!=\!0.071$).

Personality also shows a significant effect: all three archetypes share median $=0$, but differ in mean ($H(2)\!=\!84.57$, $p\!=\!4.33\times10^{-19}$, $\varepsilon^2\!=\!0.0014$). Empathetic {\agents} are the least Negative-biased ($\mu\!=\!{-0.08}$), while Analytical and Pragmatic {\agents} are more Negative-leaning ($\mu\!=\!{-0.21}$ and $\mu\!=\!{-0.25}$, respectively).

Political spectrum is statistically detectable ($U\!=\!432{,}646{,}050$, $p\!=\!4.73\times10^{-11}$, $r\!=\!0.025$), but with negligible effect size: Conservative ($\mu\!=\!{-0.23}$) and Progressive ($\mu\!=\!{-0.13}$) personas differ by only 0.10 points. This reflects the bimodal collapse: since {\agents} overwhelmingly predict extreme labels ($-2$ or $+2$), all factor levels share identical quartiles ($Q_1\!=\!{-2}$, $Q_3\!=\!{+2}$), leaving the mean as the only discriminating statistic. The statistical significance is therefore driven more by large sample size ($N=60{,}000$) than by meaningful behavioral separation.

Gender shows no measurable effect ($U\!=\!446{,}481{,}262$, $p\!=\!0.666$, $r\!=\!0.002$), with Male and Female personas indistinguishable ($\mu\!=\!{-0.18}$). 

\subsection{Persona vs.\ No-Persona: Does Conditioning Add Value?}
\label{sec:results:nopersona}

Table~\ref{tab:np_comparison} compares agreement with human ground truth across all three conditions at $\sigma\!=\!3$ (all 50 images retained), which is the most informative threshold as it includes the full image set without requiring near-unanimous human consensus.

\begin{table}[h]
  \centering
  \scriptsize
\caption{Agreement with human ground truth at $\sigma\!=\!3$. Macro F1 and Cohen's $\kappa$ ($P_2$: unweighted; $P_3$: linear; $P_5$: quadratic). No-persona uses 5 runs with modal aggregation. Confidence intervals use image-level bootstrap ($N\!=\!1{,}000$).}
  \label{tab:np_comparison}
  \resizebox{\columnwidth}{!}{
  \begin{tabular}{l|l|c|c|c|c}
    \toprule
    \multirow{2}{*}{\textbf{Task}} & \multirow{2}{*}{\textbf{Metric}} &
    \textbf{Persona} & \textbf{No-pers.} & \textbf{No-pers.} & \textbf{Largest $\Delta$ over} \\
    & & \textbf{(1,194 ann.)} & \textbf{think=T} & \textbf{think=F} & \textbf{persona baseline} \\
    \midrule
    \multirow{2}{*}{$P_2^{-}$}
      & F1      & 0.807 & \textbf{0.862} & \textbf{0.862} & $+$0.055 \\
      & $\kappa$ & 0.615 & \textbf{0.725} & \textbf{0.725} & $+$0.110 \\
    \midrule
    \multirow{2}{*}{$P_2^{+}$}
      & F1      & 0.797 & \textbf{0.836} & 0.816 & $+$0.039 \\
      & $\kappa$ & 0.597 & \textbf{0.672} & 0.634 & $+$0.075 \\
    \midrule
    \multirow{2}{*}{$P_3$}
      & F1      & 0.588 & \textbf{0.665} & \textbf{0.665} & $+$0.077 \\
      & $\kappa$ & 0.448 & \textbf{0.507} & \textbf{0.507} & $+$0.059 \\
    \midrule
    \multirow{2}{*}{$P_5$}
      & F1      & 0.309 & \textbf{0.433} & 0.425 & $+$0.124 \\
      & $\kappa$ & 0.791 & \textbf{0.879} & 0.865 & $+$0.088\\
    \bottomrule
  \end{tabular}
  }
\end{table}

The results show that the no-persona condition matches or exceeds persona-conditioned agreement across all four task variants, despite using only 5 runs per image versus $\approx$1{,}194 persona-conditioned annotations. Both think=True and think=False conditions perform similarly, with the largest improvements reaching $+$0.055 F1 on $P_2^{-}$, $+$0.077 F1 on $P_3$, and $+$0.124 F1 on $P_5$ relative to persona prompting.  The consistent outperformance across both reasoning modes confirms the result is not specific to the CoT setting.

This finding is particularly important given the asymmetry in annotation volume: the persona-conditioned system aggregates annotations from 1{,}200 agents, each evaluating an image once, yet this large-scale aggregation is matched or exceeded by only five no-persona inferences. The result suggests that the performance observed in the persona-conditioned system is attributable primarily to the MLLM's intrinsic visual understanding of the scene, rather than to any behavioral diversity introduced by persona conditioning. Persona prompting may introduce variability without improving signal in this setting.

It is important to note that the evaluation setups are not entirely symmetric: persona prompting aggregates ${\approx}1{,}194$ annotations per image, whereas no-persona uses 5 runs (5 annotations per image). Although this reduces asymmetry relative to a single run, the no-persona advantage remains consistent across both reasoning modes and all runs. A fully symmetric comparison would require substantially more no-persona repetitions and is left for future work.

\subsection{Agreement with Human Ground Truth}\label{sec:results:gt}

Table~\ref{tab:gt_agreement} reports point estimates with 95\% confidence intervals for all 12 ($\sigma$, $P$) subsets, revealing a resolution-dependent pattern of agreement: strong polarity recognition but a systematic breakdown at finer sentiment granularity.

\begin{table}[h]
    \centering
    \scriptsize
    \caption{
Agreement metrics (60\% annotation resample per image; 95\% CI via image bootstrap, $N\!=\!1{,}000$). $n_{\text{img}}$: images passing the $\sigma$ filter; $n_\text{ann}$: annotations per evaluation. Cohen's $\kappa$ is unweighted for $P_2$, linear-weighted for $P_3$, and quadratic-weighted for $P_5$. $^\dagger$: $\sigma\!=\!5$ scores of 1.000 arise from extreme filtering and should be interpreted with caution.
}
    \label{tab:gt_agreement}
    \begin{tabular}{p{1.5cm}|l|l|l|p{1.7cm}|p{1.7cm}}
        \toprule
        \textbf{Problem} & $\boldsymbol{\sigma}$ &
        $\boldsymbol{n_{\text{img}}}$ & $\boldsymbol{n_{\text{ann}}}$ &
        \textbf{Macro F1} (95\% CI) & \textbf{Cohen $\kappa$} (95\% CI) \\
        \midrule
            \multirow{3}{*}{$P_2^{-}$ (binary)}
          & 3 & 50 & 35{,}800 & 0.807 $\pm$ 0.12 & 0.615 $\pm$ 0.23 \\
          & 4 & 41 & 29{,}356 & 0.842 $\pm$ 0.12 & 0.688 $\pm$ 0.22 \\
          & 5 & 28 & 20{,}048 & 1.000$^\dagger$ $\pm$ 0.00 & 1.000$^\dagger$ $\pm$ 0.00 \\
        \midrule
        \multirow{3}{*}{$P_2^{+}$ (binary)}
          & 3 & 50 & 35{,}800 & 0.797 $\pm$ 0.11 & 0.597 $\pm$ 0.22 \\
          & 4 & 43 & 30{,}788 & 0.859 $\pm$ 0.10 & 0.720 $\pm$ 0.19 \\
          & 5 & 24 & 17{,}184 & 1.000$^\dagger$ $\pm$ 0.00 & 1.000$^\dagger$ $\pm$ 0.00 \\
        \midrule
        \multirow{3}{*}{$P_3$ (3-class)}
          & 3 & 45 & 32{,}220 & 0.588 $\pm$ 0.11 & 0.448 $\pm$ 0.21 \\
          & 4 & 34 & 24{,}344 & 0.708 $\pm$ 0.18 & 0.687 $\pm$ 0.26 \\
          & 5 & 18 & 12{,}888 & 1.000$^\dagger$ $\pm$ 0.00 & 1.000$^\dagger$ $\pm$ 0.00 \\
        \midrule
        \multirow{3}{*}{$P_5$ (ordinal-5)}
          & 3 & 36 & 25{,}776 & 0.309 $\pm$ 0.09 & 0.791 $\pm$ 0.12 \\
          & 4 & 16 & 11{,}456 & 0.514 $\pm$ 0.23 & 0.849 $\pm$ 0.17 \\
          & 5 &  4 &  2{,}864 & 1.000$^\dagger$ $\pm$ 0.00 & 1.000$^\dagger$ $\pm$ 0.00 \\
        \bottomrule
    \end{tabular}
\end{table}

The results reveal a clear \emph{polarity--granularity asymmetry}: the {\agent} reliably distinguishes positive from negative imagery, but this competence does not extend to intermediate sentiment categories. 

Binary performance ($P_2^{-}$, $P_2^{+}$) is strong from the outset—Macro F1 of 0.807 and 0.797 at $\sigma\!=\!3$, which confirms that coarse polarity discrimination is robust even without requiring strong inter-annotator consensus.

The 3-class task ($P_3$) marks the first substantive breakdown in agreement: at $\sigma\!=\!3$, F1 drops to 0.588 despite the {\agents} continuing to separate positive from negative scenes. The gap relative to binary performance is attributable almost entirely to the Neutral class, which is systematically misclassified.

The 5-class task ($P_5$) makes this pattern explicit at the most informative threshold ($\sigma = 3$), with 36 images (all classes present). Macro F1 falls to 0.309. The quadratic-weighted $\kappa$ of 0.791 at the same threshold reveals that errors are nevertheless adjacent rather than polarity inversions—ordinal direction is preserved, but intensity resolution fails, suggesting the model captures coarse ordering but not the finer distinctions required for human-like perceptual calibration. The {\agent} effectively collapses the intermediate categories (Slightly Negative, Neutral, Slightly Positive) toward the polarity extremes. This result helps explain why persona effects remain limited: the sentiment regions where persona-conditioned differences would be expected to emerge are precisely those the model compresses.

The apparent F1~$=$~1.000 at $\sigma\!=\!5$ across all tasks is an artifact of extreme filtering, not a performance ceiling. As the threshold rises, ambiguous and intermediate cases are progressively excluded: the Neutral class vanishes from $P_3$, and $P_5$ retains only four images from the polarity extremes. These perfect scores arise by construction, not by improved model capability, and should not be interpreted as evidence of strong agent capability.

\begin{figure}[htb]
    \centering
    \includegraphics[width=9cm, keepaspectratio]{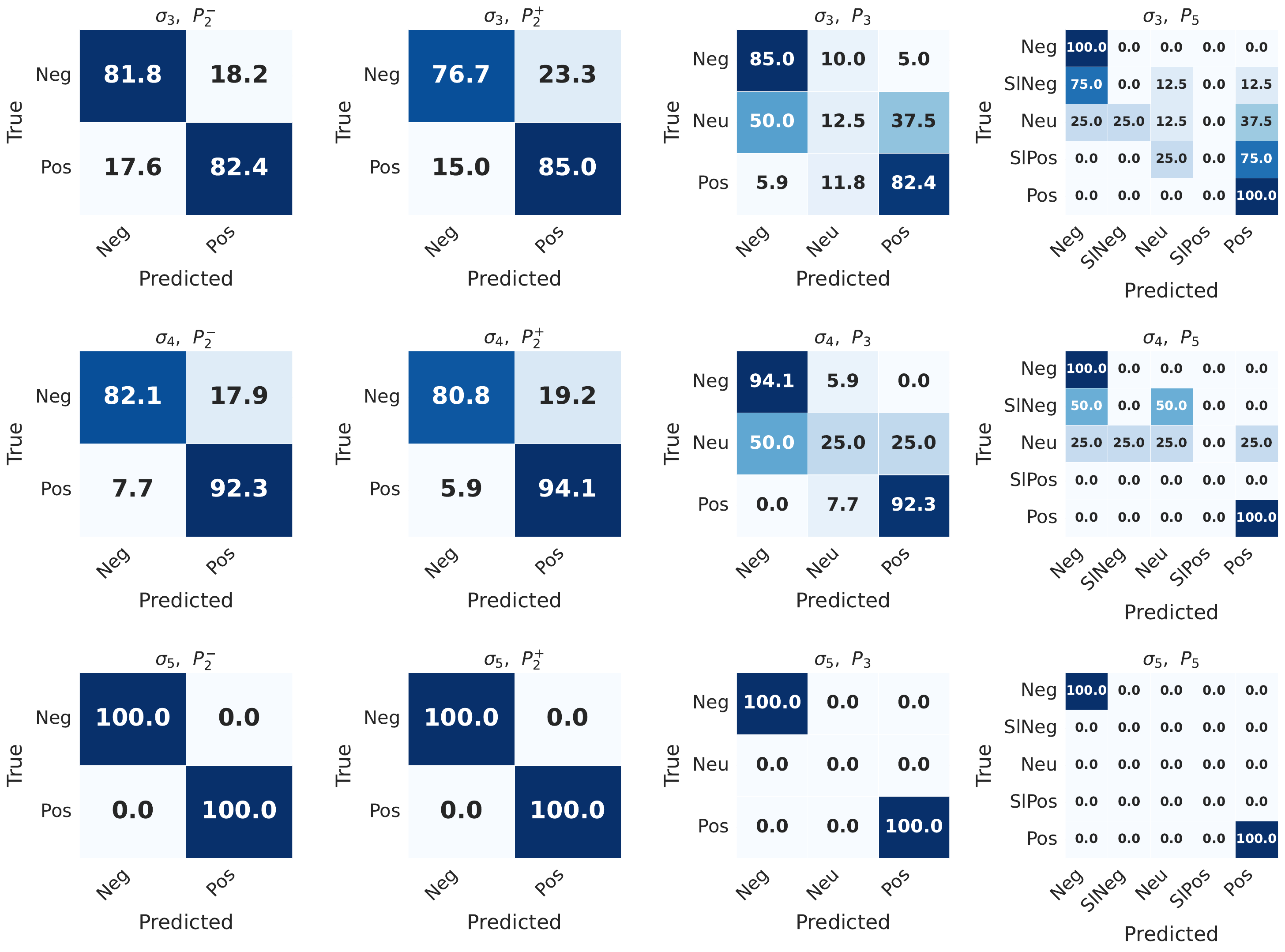}
    \caption{
        Pooled out-of-fold confusion matrices across agreement thresholds ($\sigma \in \{3, 4, 5\}$, figure rows) and problem types ($P_2^{-}$, $P_2^{+}$, $P_3$, and $P_5$, figure columns). Row-normalized. Errors in the 5-class remain adjacent labels, consistent with granularity failure rather than polarity inversion. Perfect scores at $\sigma$ = 5 arise from extreme filtering (see Section \ref{sec:results:gt}) and should not be interpreted as evidence of strong model capability.
    }
    \label{fig:confusion_baseline}
\end{figure}

Figure~\ref{fig:confusion_baseline} contextualizes these results. It shows confusion matrices across agreement thresholds ($\sigma \in \{3,4,5\}$, rows) and problem types ($P_2^{-}$, $P_2^{+}$, $P_3$, $P_5$, columns).

In binary tasks, off-diagonal mass is minimal. In the 3-class setting, Neutral images are systematically mapped to Negative or Positive. In the 5-class case, errors are confined to adjacent categories, with no polarity inversions. This has direct consequences for persona evaluation. The intermediate categories (Slightly Negative, Neutral, Slightly Positive) are precisely where human annotators disagree most and where persona-driven variation would be expected to surface. If the model systematically collapses this region, persona-induced differences in fine-grained perceptual judgment cannot be expressed, regardless of whether they exist.   

\section{Limitations and Ethical Considerations}\label{secLimiEth}

This study has several limitations. First, persona representations are defined through simple attribute labels — structured key-value injections of gender, economic status, political orientation, and personality — which may oversimplify complex social identities. The limited differentiation observed across certain attributes (e.g., gender and political orientation) may reflect this abstraction rather than the absence of real-world variation. Richer alternatives, such as biographical narratives, psychological inventories, or few-shot behavioral exemplars, may better capture the lived experiences underlying real human perspectives and remain an open direction for future work.

Second, results are based on a single multimodal LLM (\QwenVL{}) and a relatively small image sample (50 images). While the factorial design ensures controlled comparisons, broader datasets and multiple models are required to assess generalizability. Third, the observed extremity bias suggests that model outputs may not faithfully reproduce human annotation behavior, particularly for fine-grained or intermediate sentiment categories.

From an ethical perspective, the use of synthetic personas raises concerns about misrepresentation and the potential reinforcement of stereotypes. Although persona attributes are used here as experimental controls, such representations should not be interpreted as reflecting real groups. Moreover, deploying LLM-based synthetic annotators in real-world settings may risk amplifying biases if not carefully validated against human data.

These considerations highlight the need for cautious interpretation and for future work incorporating richer contextual grounding and human validation.

\section{Conclusions and Future Directions}\label{sec:conclusion}

This paper investigated whether attribute-based personas induce distinguishable and stable behavior in \agents{} performing urban image sentiment annotation. Using a controlled factorial design with 1{,}200 persona-conditioned agents across 24 persona profiles, we evaluated both within-persona convergence and cross-persona differentiation, and compared these results against a no-persona ablation.

Four main findings emerge. First, persona prompts produce strong within-persona convergence, indicating stable and reproducible judgments. Second, cross-persona differentiation is limited: economic status and personality induce statistically detectable but practically modest variation, while gender and political orientation contribute little meaningful separation. Third, the dominant failure mode concerns sentiment granularity rather than polarity discrimination, with agents collapsing intermediate categories despite preserving coarse ordinal structure. Fourth, despite the substantial annotation volume invested in persona conditioning, a five-run modal no-persona prediction matched or exceeded persona-conditioned agreement, suggesting the benefit of persona conditioning in this setting warrants further investigation — though reasoning mode and annotation volume asymmetry should be controlled in future work before drawing definitive conclusions.

Taken together, these findings reveal an important resolution-dependent limitation of simple attribute-based persona prompting: while it supports behavioral stabilization and coarse-grained polarity judgments, it does not generate sufficiently rich variation to capture fine-grained perceptual judgments characteristic of human annotators.

Several directions follow. First, richer persona representations grounded in social context and biographical narratives may better elicit meaningful variation. Second, prompt interventions aimed at mitigating extremity bias may improve calibration for intermediate sentiment categories. Third, broader validation across models, prompting strategies, and datasets is needed, including further study of whether persona prompting contributes diversity beyond accuracy.

More broadly, these findings raise questions about the adequacy of simple attribute-based personas as proxies for human diversity, while suggesting that their value may lie more in stabilizing behavior than in generating rich perceptual variation.

\section{Acknowledgments}

This study is partially supported by the National Council for Scientific and Technological Development - CNPq (processes 314603/2023-9, 441444/2023-7, and 444724/2024-9) and INCT TILD-IAR (proc. 408490/2024-1).

\bibliographystyle{ieeetr}
\bibliography{references}

\end{document}

%% file: diagrams/methodology_overview_v2.tex
\definecolor{mopal1}{HTML}{FEEAC9}
\definecolor{mopal2}{HTML}{FFCDC9}
\definecolor{mopal4}{HTML}{FD7979}

\begin{tikzpicture}[scale=0.78, every node/.style={scale=0.78},
    box/.style  = {draw=black!80, rounded corners=5pt, minimum width=30mm,
                   minimum height=14mm, font=\small\sffamily\bfseries, align=center,
                   inner sep=5pt, text width=30mm, text=black!80},
    dashbox/.style = {draw=black!55, rounded corners=5pt, dashed, minimum width=30mm,
                   minimum height=14mm, font=\small\sffamily\bfseries, align=center,
                   inner sep=5pt, text width=30mm, text=black!55},
    oval/.style = {draw=black!80, ellipse, minimum width=28mm,
                   minimum height=12mm, font=\small\sffamily\bfseries, align=center,
                   text width=26mm, text=black!80},
    arr/.style  = {-{Stealth[length=5pt]}, thick, black!80},
    darr/.style = {-{Stealth[length=5pt]}, thick, black!45, dashed},
  ]

  \node[box, fill=mopal1] at (-0.1, 0) (phase1)
        {1{,}200 Persona-Driven\\Agents\\[2pt]{\scriptsize 24 profiles $\times$ 50 agents}\\{\scriptsize Balanced full factorial}};

  \node[box, fill=mopal2] at (3.6, 1.5) (annot)
        {Persona\\Annotation\\[2pt]{\scriptsize Qwen3-VL:8B \textbf{think=T}}\\{\scriptsize 50 urban images}};

  \node[dashbox, fill=mopal1!35] at (3.6, -1.5) (ablat)
        {No-Persona\\Ablation\\[2pt]{\scriptsize Qwen3-VL:8B}\\{\scriptsize think=T / think=F}};

  \node[oval, fill=mopal4, text=white] at (7.6, 0) (eval)
        {Convergence \&\\Agreement\\Analysis};

  \draw[arr]  (phase1.east) |- (annot.west);
  \draw[darr] (phase1.east) |- (ablat.west);
  \draw[arr]  (annot.east)  -| (eval.north);
  \draw[darr] (ablat.east)  -| (eval.south);

  \node[font=\small\sffamily\bfseries, text=black!80] at (0,   2.8)
        {Phase~1: Persona Design};
  \node[font=\small\sffamily\bfseries, text=black!80] at (3.8, 2.8)
        {Phase~2: Annotation};
  \node[font=\small\sffamily\bfseries, text=black!80] at (7.6, 2.8)
        {Phase~3: Evaluation};

  \node[font=\scriptsize\sffamily, text=black!80, align=left] at (7.6, -2.5) {%
    \rule[0.5ex]{6mm}{0.6pt}\ main \quad%
    \begin{tikzpicture}[baseline=-0.5ex]%
      \draw[darr, shorten >=0pt] (0,0) -- (6mm,0);%
    \end{tikzpicture}\ ablation%
  };

\end{tikzpicture}

%% file: diagrams/annotation_pipeline_redopt.tex
\definecolor{pal1}{HTML}{FEEAC9}
\definecolor{pal2}{HTML}{FFCDC9}
\definecolor{pal3}{HTML}{FDACAC}
\definecolor{pal4}{HTML}{FD7979}

\begin{tikzpicture}[
    box/.style  = {draw=black!80, rounded corners=4pt, minimum width=18mm,
                   minimum height=7mm, font=\tiny\sffamily, align=center,
                   inner sep=-1pt, text width=18mm, text=black!80},
    doc/.style  = {draw=black!80, chamfered rectangle, minimum width=18mm,
                   minimum height=7mm, font=\tiny\sffamily, align=center,
                   inner sep=0pt, text width=18mm, text=black!80},
    oval/.style = {draw=black!80, ellipse, minimum width=12mm,
                   minimum height=5mm, font=\tiny\sffamily\bfseries,
                   text=black!80},
    arr/.style  = {-{Stealth[length=3.5pt]}, thick, black!80},
    darr/.style = {-{Stealth[length=3.5pt]}, thick, dashed, black!80},
    lbl/.style  = {font=\tiny\sffamily\bfseries, text=black!80, align=center, fill=white, inner sep=1pt},
  ]
  
  \node[oval, fill=pal1] at (0, 0)    (start) {START};

  \node[box, fill=pal1] at (0, -1.0)  (loader)   {Image Loader};
  \node[box, fill=pal1] at (0, -2.0)  (assembler){Prompt Assembler};

  \node[box, fill=pal2] at (0, -3.6)  (annot)
        {Vision Annotator\\[1pt]{\tiny Qwen3-VL:8B}\\{\tiny T=0.1, think=T}};

  \node[box, fill=pal3] at (0, -5.2)  (router)   {Evaluate\\Response};

  \node[box, fill=pal2] at (-3.2, -5.2) (ftout)  {Compact\\Retry};
  
  \node[box, fill=pal2] at (3.2, -5.2)  (flabel) {Fuzzy\\Normalise};
  \node[box, fill=pal2] at (3.2, -6.6)  (fjson)  {Regex\\Extract};

  \node[doc, fill=pal4, text=white] at (-1.3, -7.2) (out)
        {Output\\[1pt]{\tiny Image sentiment (JSON)}};
  \node[doc, fill=pal3] at (1.3, -7.2) (fail)
        {Log\\[1pt]{\tiny annotation of failures}\\{\tiny (JSONL)}};

  \node[oval, fill=pal1] at (0, -8.6) (end)   {END};

  \draw[arr] (start.south)     -- (loader.north);
  \draw[arr] (loader.south)    -- (assembler.north);
  \draw[arr] (assembler.south) -- (annot.north);
  \draw[arr] (annot.south)     -- (router.north);

  \draw[darr] (router.west) -- node[lbl, above] {\tiny 360s Timeout} (ftout.east);
  
  \draw[darr] (router.east) -- node[lbl, above] {\tiny Bad Label} (flabel.west);
  \draw[darr] ([yshift=-4pt]router.east) to[out=0, in=180] node[lbl, sloped, above, pos=0.6] {\tiny JSON Fail} (fjson.west);

  \draw[arr] (ftout.west) -- ++(-0.4,0) |- (annot.west);
  
  \draw[arr] (flabel.east) -- ++(0.4,0) |- ([yshift=-4pt]annot.east); 
  \draw[arr] (fjson.east)  -- ++(0.8,0) |- ([yshift=4pt]annot.east);  

  \draw[arr] (router.south) to[out=270, in=90] node[lbl, left=2pt] {\tiny Success} (out.north);
  \draw[arr] (router.south) to[out=270, in=90] node[lbl, right=2pt] {\tiny Exhausted} (fail.north);

  \draw[arr] (out.south)  to[out=270, in=180] (end.west);
  \draw[arr] (fail.south) to[out=270, in=0]   (end.east);

\end{tikzpicture}